\documentclass[letterpaper, 10 pt, conference]{ieeeconf}  

\IEEEoverridecommandlockouts                              

\let\labelindent\relax
\usepackage{times}
\usepackage{epsfig}
\usepackage{graphicx}
\usepackage{amsmath}
\usepackage{amssymb}
\usepackage{verbatim}
\usepackage{mathtools}
\usepackage{enumitem}
\usepackage{booktabs}
\usepackage{xcolor}
\usepackage{multirow}
\usepackage{subcaption}
\usepackage{algorithm} 
\usepackage{algpseudocode} 
\usepackage{pifont}
\newcommand{\cmark}{\ding{51}}%
\newcommand{\xmark}{\ding{55}}%
\newcommand{\algrule}[1][.2pt]{\par\vskip.5\baselineskip\hrule height #1\par\vskip.5\baselineskip}

\makeatletter
\let\NAT@parse\undefined
\newcommand\fs@betterruled{%
  \def\@fs@cfont{\bfseries}\let\@fs@capt\floatc@ruled
  \def\@fs@pre{\vspace*{5pt}\hrule height.8pt depth0pt \kern2pt}%
  \def\@fs@post{\kern2pt\hrule\relax}%
  \def\@fs@mid{\kern2pt\hrule\kern2pt}%
  \let\@fs@iftopcapt\iftrue}
\floatstyle{betterruled}
\restylefloat{algorithm}
\makeatother
\usepackage[pagebackref=true,breaklinks=true,letterpaper=true,colorlinks,citecolor=blue,linkcolor=blue,bookmarks=false]{hyperref}

\title{\LARGE \bf
Who Make Drivers Stop? Towards Driver-centric Risk Assessment: Risk Object Identification via Causal Inference
}

\author{Chengxi Li$^{1}$ \, Stanley H. Chan$^{1}$ \, Yi-Ting Chen$^{2}$
\thanks{C. Li$^{1}$ and S. H. Chan$^{1}$ are with Department of Electrical and Computer Engineering, Purdue University, West Lafayette, IN, USA. Y.-T. Chen$^{2}$ is with Honda Research Institute USA, San Jose, CA, USA.}%
\thanks{The work was done when C. Li was an intern at Honda Research Institute, USA.}
}

\begin{document}

\maketitle
\thispagestyle{empty}
\pagestyle{empty}

\begin{abstract}
A significant amount of people die in road accidents due to driver errors.
To reduce fatalities, developing intelligent driving systems assisting drivers to identify potential risks is in an urgent need.
Risky situations are generally defined based on collision prediction in the existing works.
However, collision is only a source of potential risks, and a more generic definition is required.
In this work, we propose a novel driver-centric definition of risk, i.e., objects influencing drivers’ behavior are risky.
A new task called risk object identification is introduced.
We formulate the task as the cause-effect problem and present a novel two-stage risk object identification framework based on causal inference with the proposed object-level manipulable driving model. 
We demonstrate favorable performance on risk object identification compared with strong baselines on the Honda Research Institute Driving Dataset (HDD).
Our framework achieves a substantial average performance boost over a strong baseline by 7.5\%.
\end{abstract}
\section{INTRODUCTION}
More than 1.3 million people die in road accidents worldwide every year, approximately 3,700 people per day~\cite{who}.
Car accident deaths are the $1^{st}$ leading causes of death, excluding health illness.
A massive number of car accident fatalities are due to driver errors such as the lack of awareness~\cite{nhtsa}.
To reduce the fatality rate, developing intelligent driving systems assisting drivers to identify potential risks is in an urgent need.
%
The identification of potential risks has been studied extensively in the risk assessment literature~\cite{risk_assessment_Lefevre_ROBOMECH}.
At the core of risk assessment is the definition of risk.
In the context of intelligent vehicles, the risk is generally defined based on collision prediction.
While the definition is widely used, road collision is only a source of potential hazards in driving~\cite{risk_assessment_Lefevre_ROBOMECH}.

\begin{figure}[t!]
\includegraphics[width = \columnwidth]{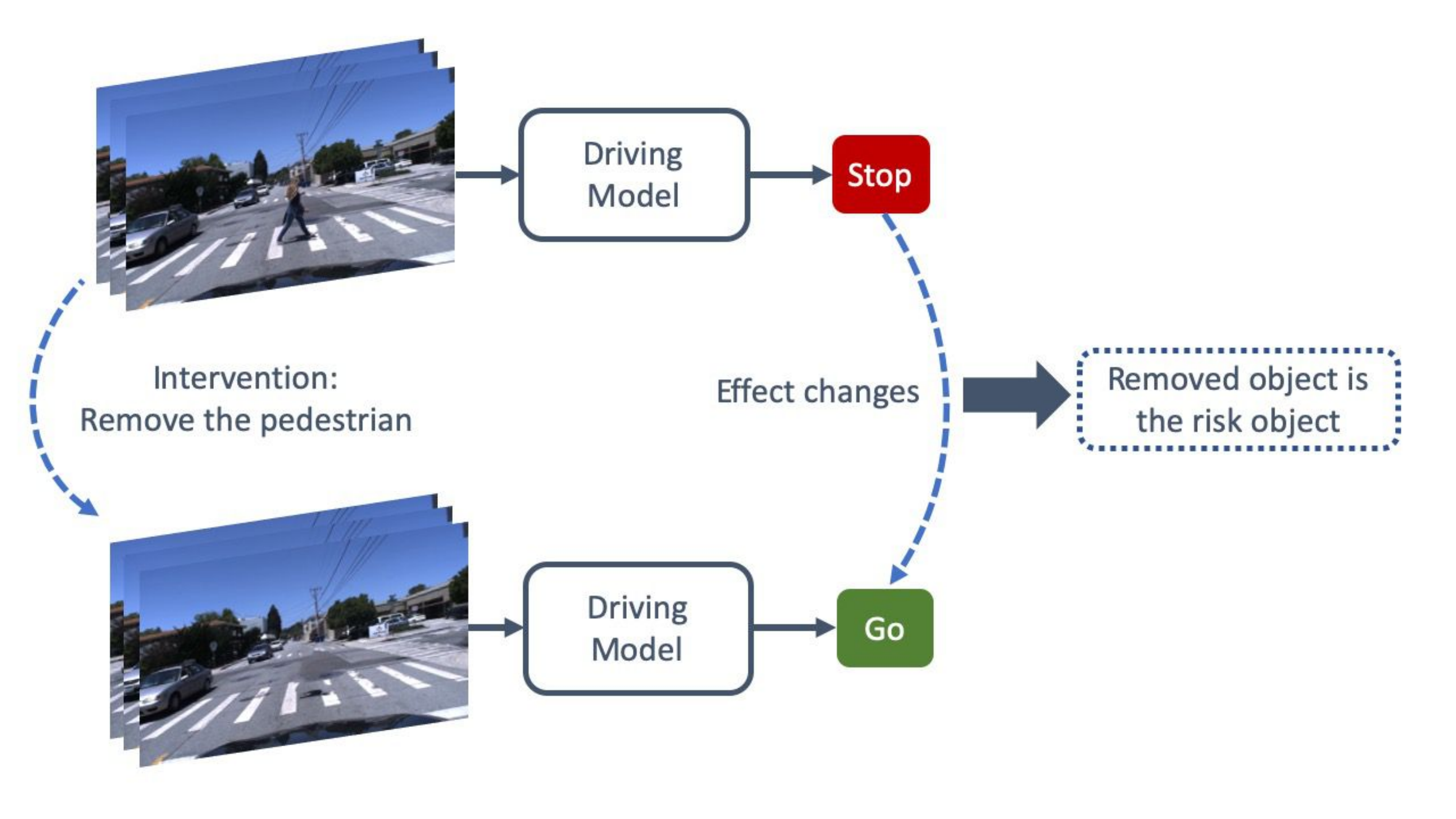}
\vspace{-18pt}
\caption{A two-stage risk object identification framework. Intervening the input by removing the pedestrian changes the driver's behavior (effect) from `Stop' to `Go', indicating the removed object is the risk object (cause) for `Stop'.}
\label{fig:illustration}
\end{figure}

%
%
In this paper, we propose a driver-centric definition of risk, i.e., \textit{objects influencing drivers’ behavior are risky}.
%
%
Imagine driving in a scenario shown in Fig.~\ref{fig:illustration} that we are planning to pass the intersection while yielding to a crossing pedestrian.
Without the presence of the pedestrian, we would have passed the intersection without stopping.
In other words, if we did not slow down or stop, a dangerous situation would occur.
During our daily driving, we frequently interact with different traffic participants under diverse configurations.
These reactive scenarios are commonly encountered than collisions.
Thus, we believe that the proposed definition captures various learning signals for assessing risk. 
%
%

A natural question arises: \textit{who makes drivers stop?}
%
%
We propose a new task called \textit{risk object identification}, which aims to identify the object influencing drivers' behavior.
%
The proposed task can be approached via two existing methodologies: (1) supervised learning algorithms that learn to localize risky regions~\cite{Zeng_risk_cvpr2017}, recognize important objects~\cite{Gao_goal_icra2019}, and imitate drivers' gaze behavior~\cite{Alletto_Dreye_cvprw2016}, and (2) salient regions/objects identification via self-attention mechanisms in end-to-end networks~\cite{kim2017interpretable,wang2018deep}.
%

In the first category, risky regions and important objects are identified by formulating as a two-class object detection problem.
Specifically, object-object interaction~\cite{Zeng_risk_cvpr2017,Gao_goal_icra2019}, object-environment interaction~\cite{Zeng_risk_cvpr2017}, object motion trajectory~\cite{Zeng_risk_cvpr2017,Gao_goal_icra2019} are designed to enable models to identify risky regions or important objects.
Alternatively, objects influencing drivers' behavior can be obtained via predicting pixel-level driver’s attention learned by imitating human gaze behavior~\cite{Alletto_Dreye_cvprw2016,Xia_ACCV_2018,Tawari_saliency_itsc2018}.
While promising performances have been shown, labeling risky regions or important objects for training two-class object detectors requires extensive human-labeled annotations.
%
Human gaze behavior is intrinsically noisy, and fixations may not directly associate with objects influencing drivers' behavior.  

For the second strategy, objects influencing behaviors of drivers can be formulated as selecting regions/objects with high activations in visual attention heat maps learned from end-to-end driving models~\cite{kim2017interpretable,wang2018deep}.
Kim and Canny~\cite{kim2017interpretable} proposed a two-stage framework.
In the first stage, a visual attention mechanism is designed to highlight image regions that influence the end-to-end driving model's output.
%
%
A causal filtering step as the second stage is applied to determine the regions influencing the network's behavior.
%
%
%
Wang et al.~\cite{wang2018deep} incorporated an object-level attention mechanism into end-to-end driving models to increase robustness and interpretability.
%
%
Both attention mechanisms are trained to optimize task-dependent objective functions.
However, there is no guarantee that networks will attend the regions/objects that influence drivers' behavior.
It is worth noting that the issue `causal misidentification' is discussed in training end-to-end driving models~\cite{Haan_causalconfusion_nips2019}.

To address the issues mentioned above, we formulate the definition as the cause-effect problem~\cite{Pearl_causality_2009} and propose a novel risk object identification framework.
The core concept is depicted in Fig.~\ref{fig:illustration}.
First, an object-level manipulable driving model is learned to predict drivers’ behavior. In this work, we simplify the possible driver behaviors to be `Go' or `Stop.' Note that modeling of more fine-grained drivers' behavior is our future work.
%
%
%
%
Second, given a `Stop' prediction (i.e., driver behavior is influenced by objects), we intervene input video by removing a tracklet at a time and inpainting the removed area in each frame to simulate a scenario without the presence of the tracklet.
The trained driving model is used to predict the corresponding driver behavior.
Note that we assume that the cause of driver behavioral change is vehicles or pedestrians in this work.
%
%
The object causes the maximum effect change is the one that influences drivers’ behavior.
%
We benchmark the proposed framework on the Honda Research Institute Driving Dataset (HDD)~\cite{Ramanishka_behavior_CVPR_2018}.
%
%
%
Experimental results show that the proposed framework achieves favorable risk object identification performance compared with strong baselines, both quantitatively and qualitatively.
Furthermore, extensive ablation studies are conducted to justify our design choices.

The contributions of this work are summarized as follows.
First, we propose a novel driver-centric definition of risk, i.e., objects influencing drivers’ behavior are risky, and a new task called risk object identification is introduced.
Second, we formulate the task as the cause-effect problem and propose a novel two-stage risk object identification framework based on causal inference with the proposed object-level manipulable driving model.
Third, we demonstrate favorable performance on risk object identification in comparison with strong baselines on the HDD dataset~\cite{Ramanishka_behavior_CVPR_2018}.


\section{Related Work}
\subsection{Risk Assessment}
Living agents can assess risk for decision making. Lef{\`e}vre et al.~\cite{risk_assessment_Lefevre_ROBOMECH} surveyed existing methods for motion prediction and risk assessment in the context of intelligent vehicles.
A popular risk assessment methodology is to predict all possible colliding future trajectories.
%
%
While many works follow the direction, Lef{\`e}vre et al.~\cite{Lefevre_iros2012} defined the computation of risk as to the probability that expectation and intention do not match.
The paradigm is very closed to the proposed definition of risk, i.e., objects influencing drivers' behavior, the underlying risk object identification process is different.
In~\cite{Lefevre_iros2012}, a risk object is identified by computing the `hazard probability' based on the same computation of risk.
%
We recognize the risk object by simulating the causal effect by removing an object using end-to-end driving models.
%


\subsection{Causal Confusion in End-to-end Driving Models}
Recent successes~\cite{Bojarski_nvidia_2016,Xu_drivingmodel_CVPR_2017} demonstrated that a driving policy can be learned in a supervised manner from human demonstration~\cite{Codevilla_ICRA_2018,wang2018deep,Kim_advice_CVPR_2019}.
Additionally, recent driving datasets~\cite{Ramanishka_behavior_CVPR_2018,Yu_bdd100k_arXiv2018} with high-quality drivers' demonstration, enable training driving models under different real-world scenarios.
However, the issue of causal confusion in training end-to-end driving models is raised in~\cite{Codevilla_imitation_arXiv2019,Hawke_CIL_arXiv2019,Haan_causalconfusion_nips2019}.
Haan et al.~\cite{Haan_causalconfusion_nips2019} proposed incorporating the concept of functional causal models~\cite{Pearl_causality_2009} into imitation learning to address the issue of `causal misidentification.'
In~\cite{Hawke_CIL_arXiv2019}, they overcame the causal misidentification issue by adding noises to inputs.

Our work is complementary to~\cite{Hawke_CIL_arXiv2019,Haan_causalconfusion_nips2019}.
Specifically, the focus of~\cite{Hawke_CIL_arXiv2019,Haan_causalconfusion_nips2019} is to improve the robustness of driving models.
Instead, our proposed driving model is to enable intervention for risk object identification.
We believe the two lines of work should be studied jointly to obtain robust driving models with explicit reasoning mechanisms.


\begin{figure*}[h!]
\includegraphics[width = \textwidth]{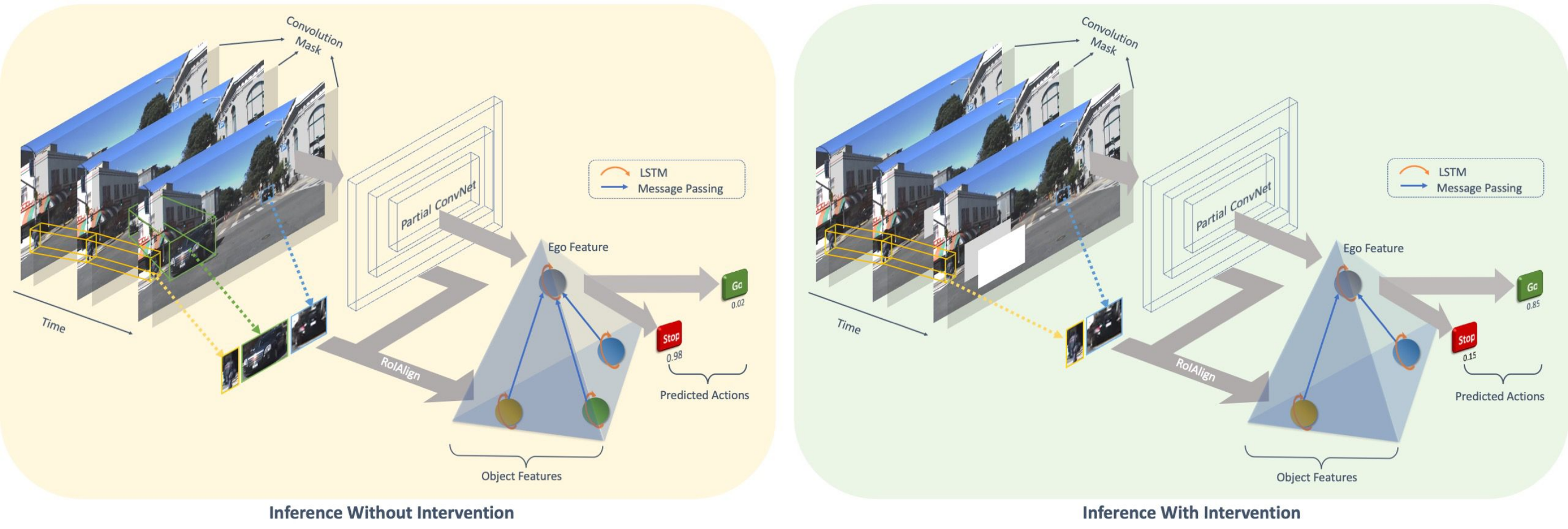}
    \caption{An overview of our framework. The right and left figures show the inference process with and without intervention, respectively. Both employ the same driving model to output the predicted driver behavior. The inputs to the driving model include a sequence of RGB frames, a sequence of binary masks, and object tracklets. Partial convolution and average pooling are employed to obtain the ego features, while object features are extracted by RoIAlign. Each feature is modeled temporally and then propagates information to form a scene representation for final prediction. On the right, the input is intervened at an object level by masking the selected object on the convolution mask and removing it from the tracklets. For example, we remove the car in the green box, and the driving model returns a high confidence score of `Go.'}
    \label{fig:framework}
\end{figure*}

\section{Method}

We formulate the risk object identification problem as the cause-effect problem~\cite{Pearl_causality_2009}. 
Specifically, we leverage and realize the concept of causal inference in a two-stage framework to identify the cause (i.e., the object) of an effect (i.e., driver behavioral change) via the proposed object-level manipulable driving model. 
We discuss the methodology of the proposed framework in the following.

\subsection{Object-level Manipulable Driving Model}

%
%
To realize causal inference for risk object identification, we propose an object-level manipulable driving model with the two properties.
First, the driving model should be able to predict the corresponding driver behavior in an intervened scenario, i.e., an object is removed from historical observations.
Second, the driving model needs to predict driver behavior while interacting with traffic participants.
To realize the properties, we use \textit{partial convolutional layers}~\cite{liu2018image} instead of standard convolutional layers.
%
A partial convolutional layer is initially introduced for image inpainting.
We utilize partial convolutions to simulate a scenario without the presence of an object.
%
%
%
A partial convolutional layer takes two inputs, i.e., an RGB frame and corresponding one-channel binary mask.
The value of pixels within a mask is set to be 1 by default.
While training the driving model with an intervention (Section~\ref{section:training}) and perform causal inference for risk object identification (Section~\ref{section:inference}), we set the pixels within as elected object to be 0.

To obtain representations of objects, Mask R-CNN~\cite{HeCVPR2017}, and Deep SORT~\cite{wojke2017simple} is applied to detect and track every object.
RoIAlign~\cite{HeCVPR2017} is employed to extract the corresponding object representation.
The ego's representation is extracted via average pooling the features extracted from the partial convolution networks.
Here we use `ego' to denote the representation of a driver.
%
We use a long short-term memory(LSTM) module~\cite{hochreiter1997long} to model the ego and objects' temporal dynamics.

Motivated by~\cite{wang2018deep} and \cite{Chengxi2019}, modeling interaction is essential for driver behavior modeling.
We model interactions between the ego and objects via the following message passing mechanism,
\begin{equation}
    g = h_e \oplus \frac{1}{N}(\sum_{i=1}^{N} h_i)
\end{equation}
where $g$ is defined as the aggregated representation. The ego's representation $h_e$ is obtained after temporal modeling and $h_o = \{h_i, h_2, \cdots, h_N\}$ are representations of $N$ objects.
$\oplus$ indicates a concatenation operation.
To manipulate the representation at the object level, we set the pixel value of the binary mask to 0 at the selected object's location.
The mask influences representations extracted from the partial convolution networks and disconnect the message of the selected object from the rest of the objects and the ego.
This representation $g$ is passed through fully connected layers to obtain the final classification of driver behavior.
In this work, we categorize driver behavior to be `Go' and `Stop'.
An overview of the proposed driving model architecture is shown in Fig. \ref{fig:framework}.

\subsection{Training with Intervention}\label{section:training}

\begin{algorithm}[t]

	\caption{Training Driving Model with Intervention}
	\begin{algorithmic}
	    \State $T$: Number of frames 
    	\State $N$: Number of objects in the given tracklet list
	    \State $h_e$: Hidden states of ego in LSTM module
	    \State $h_o$: Hidden states of objects in LSTM Module $h_o \coloneqq \{h_1, h_2, \cdots, h_N\}$
	    \State \textbf{A}: Ground truth driver behavior (either `GO' or `STOP')
	    \State \textbf{Input}: A sequence of RGB frames $I \coloneqq \{I_1, I_2,\cdots, I_T\}$
        \State \textbf{Output}: Confidence score of driver behaviors $s^{go}, s^{stop}$ 
	    \algrule
	\end{algorithmic}
		\begin{algorithmic}[1]
	    \State $O 
         \begin{aligned}[t]
            &  \coloneqq\texttt{DetectionAndTracking($I$) } \\
            & \coloneqq \{O_1, O_2, \cdots, O_N\}   \text{ // List of tracklets } 
         \end{aligned}$  

        
        \If{\textbf{A} is `GO' and $N > 1$}
            \State $\begin{aligned}[t] &\text{// Randomly choose one object to remove} \\
            & k \coloneqq \texttt{RandomSelect}(N)
            \end{aligned}$
        \Else
            \State $k$ is empty
        \EndIf
            
        \State $\begin{aligned}[t] &\text{// Mask out the region of object $k$ on each mask frame} \\ 
            & M \coloneqq \texttt{MaskGenerator}(I, O_k)
            \end{aligned}$ 
        
        \State $ \begin{aligned}[t] &\text{// Remove the object $k$ from the tracklet list} \\
        & O' = O - \{O_k\}
        \end{aligned}$ 
        
        \State $s^{go}, s^{stop} \coloneqq \texttt{DrivingModel}(I,M,O')$ //Defined as below
        \State \Return $s^{go}, s^{stop}$
            \algrule
	\end{algorithmic} 
		\begin{algorithmic}[1]
	\Function{DrivingModel}{$I, M, O$}
        \For{$t \in \{1,2,\cdots, T \}$}
            \State $e_t \coloneqq \texttt{EgoFeature}(I_t, M_t)$
            \State $h_e \coloneqq \texttt{LSTM} (e_t, h_e)$
            \For{$ O_i \in O $}
                \State $f_i^t \coloneqq \texttt{RoIAlign}(o_i^t)$ // Object Features ($o_i^t$ is $i$-th object's bounding box at time $t$)
                \State $h_i \coloneqq \texttt{LSTM}(f_i^t, h_i)$ 
            \EndFor
        \EndFor
        \State $g \coloneqq \texttt{MessagePassing}(h_e,h_o)$
        \State $s^{go}, s^{stop} \coloneqq \texttt{ActionClassifier}(g)$
        \State \Return $s^{go}, s^{stop}$
    \EndFunction
	\end{algorithmic} 
	\label{algo1}
\end{algorithm} 

We cluster and label the training samples into two categories — (1) `Go': the ego vehicle moves without stopping/yielding/deviating and (2)`Stop': the ego vehicle stops, yields, or deviates for objects.
We use the behavioral change between these two states to reason the risk object that influences driver behavior.
It is worth noting that this is the only supervision signal.
%
The performance of driving models can be improved by training with samples with different traffic configurations~\cite{Xu_drivingmodel_CVPR_2017}.
%
Due to the limited real-world human driver demonstrations, we design a training strategy that utilizes the concept of intervention from causal inference~\cite{Pearl_causality_2009} to improve the robustness of the driving model.
Specifically, we create new configurations based on a simple yet effective notion, i.e., removing non-causal objects does not affect driver behavior.
For instance, in the `Go' scenario, the ego vehicle goes straight and passes an intersection while pedestrians on the sidewalk.
It is reasonable to assume if a pedestrian was not present, the behavior of the ego vehicle is the same.

This strategy is only applicable to the first category.
%
In the second category, we need to know the causal object to remove non-casual objects.
Intensive labeling of risk objects' locations is required.
Note that this contradicts the spirit of the proposed weakly supervised method. 
%
%
Moreover, even if the annotations of causal objects are given, we cannot intervene on the causal object and simply assume the corresponding driver behavior to `Go' because traffic situations are inherently complicated, making the intervened diver behavior unclear.
For instance, imagine that under a congestion circumstance where the ego vehicle stops for the front vehicle at an intersection, while the traffic light shows red.
In such a case, the front vehicle is labeled as the risk object (cause).
However, removing the front vehicle does not necessarily change the ego vehicle's behavior because of the red light.

Algorithm~\ref{algo1} provides the pseudo-code of the proposed training process.
For training samples in the first category, we randomly select one object $k$ to intervene.
%
Based on detection and tracking, a one-channel binary mask is generated and set the pixels' value within the $k$-th object's region to be 0.
\begin{equation}
 M_t(i, j) =
\begin{cases}
    0,& \text{if } (i, j) \text{ in region } o_k^t\\
    1,& \text{otherwise}
\end{cases}
\end{equation}
where $M_t$ denotes the generated mask at time $t$, $o_k^t$ is the bounding box of the $k$-th object at time $t$, and $(i,j)$ is the pixel coordinate.
This mask and the corresponding RGB frame will be the inputs to the partial convolutional networks.
Notice that the $k$-th object is discarded from the tracklet list before feeding into the driving model.

\subsection{Causal Inference for Risk Object Identification}\label{section:inference}
\begin{algorithm}[t]
\caption{Inference for Risk Object Identification}
\begin{algorithmic}
   \State $T$: Number of frames
    \State $N$: Number of objects
   \State \textbf{Input}: A sequence of RGB frames $I \coloneqq \{I_1, I_2,\cdots, I_T\}$ where the ego car stops
        \State \textbf{Output}: Risk object ID
   \algrule
\end{algorithmic}
\begin{algorithmic}[1]
   \State $O
         \begin{aligned}[t]
            &  \coloneqq\texttt{DetectionAndTracking($I$) } \\
            & \coloneqq \{O_1, O_2, \cdots, O_N\} \text{// List of tracklets }
         \end{aligned}$  

        \For{$ O_k \in O $}
            \State $\begin{aligned}[t] &\text{// Mask out the region of object $k$ on each frame} \\
            & M \coloneqq \texttt{MaskGenerator}(I, O_k)
            \end{aligned}$
             \State $\begin{aligned}[t] &\text{// Remove the object $k$ from the tracklet list}\\
             & O' = O - \{O_k\}
             \end{aligned}$
            \State $\begin{aligned}[t] & \text{// Predict the action of ego car without object $k$ }\\
            & s^{go}_k, s^{stop}_k  \coloneqq \texttt{DrivingModel}(I,M, O')
             \end{aligned}$
        \EndFor
   
    \State \Return $\arg\max \limits_{k}(s^{go}_k) $
\end{algorithmic}
\label{algo3}
\end{algorithm}

%
The concept of causal inference is utilized for risk object identification.
Specifically, given video frames and tracklets, the masks of a tracklet and the corresponding video frames are processed by the trained driving model.
The driving model predicts the confidence score of `Go' and `Stop.'
%
%
We select the object with the highest confidence score of `Go', indicating this object causes the most substantial driver behavioral change as the risk object.
Algorithm~\ref{algo3} describes the process.

\begin{table}[t]
    \small
    \centering
    \vspace{5pt}
    \resizebox{\columnwidth}{!}{
        
        \begin{tabular}
            {@{}cc@{\;}@{\;}c@{\;}@{\;}c@{\;}@{\;}c@{}}
            
        \toprule
        & \begin{tabular}{@{}c@{}}\# Training Frames \end{tabular}
        & \begin{tabular}{@{}c@{}} \# Validation Frames \end{tabular}
        & \begin{tabular}{@{}c@{}} \# Test Frames \end{tabular}\\

        \cmidrule{1-4}
        Crossing Vehicle & 18,696 & 5,652 & 311  \\
        Crossing Pedestrian & 20,784 & 3,999 & 84   \\
        Parked Vehicle & 11,484 & 3,537 & 136   \\
        Congestion& 21,132 & 5,607 & 99  \\

    \bottomrule
    \end{tabular}
    }

    \caption{Statistics of train/val/test samples in HDD~\cite{Ramanishka_behavior_CVPR_2018} used in our experiments.}
    \label{table:dataset_statistics}
    \vspace{-8pt}
\end{table}

\begin{table*}[t]
    \small
    \centering
    \vspace{5pt}
    \resizebox{\textwidth}{!}{
        
        \begin{tabular}
            {@{} l  c @{\;}  c  @{\;} 
            @{\;} c @{\;}   @{\;} c @{\;}    @{\;} c  @{\;}
            @{\;} c @{\;}   @{\;} c @{\;}    @{\;} c  @{\;}
            @{\;} c @{\;}   @{\;} c @{\;}    @{\;} c  @{\;} 
            @{\;} c @{\;}   @{\;} c @{\;}    @{\;} c  @{}}
            
        \toprule
         \multirow{3}{*}{ \begin{tabular}{@{}c@{}}Driving \\ Model\end{tabular}}  &  \multirow{3}{*}{Mask} &  \multirow{3}{*}{ \begin{tabular}{@{}c@{}} Training with \\ Intervention \end{tabular} } & \multicolumn{3}{c}{Crossing Vehicle} &   \multicolumn{3}{c}{Crossing Pedestrian} & \multicolumn{3}{c}{Parked Vehicle} & \multicolumn{3}{c}{Congestion}  \\
        \cmidrule(lr){4-6} \cmidrule(lr){7-9}\cmidrule(lr){10-12}\cmidrule(lr){13-15}
             &
             &
             &
            \begin{tabular}{@{}c@{}} $Acc_{0.5}$ \end{tabular} &
            \begin{tabular}{@{}c@{}} $Acc_{0.75}$\end{tabular} &
            \begin{tabular}{@{}c@{}}$mAcc$\end{tabular} &
            \begin{tabular}{@{}c@{}}$Acc_{0.5}$\end{tabular} &
            \begin{tabular}{@{}c@{}} $Acc_{0.75}$\end{tabular} &
            \begin{tabular}{@{}c@{}} $mAcc$\end{tabular} &
            \begin{tabular}{@{}c@{}}$Acc_{0.5}$\end{tabular} &
            \begin{tabular}{@{}c@{}}$Acc_{0.75}$\end{tabular} &
            \begin{tabular}{@{}c@{}}$mAcc$\end{tabular} &
            \begin{tabular}{@{}c@{}} $Acc_{0.5}$\end{tabular} &
            \begin{tabular}{@{}c@{}}$Acc_{0.75}$\end{tabular} &
            \begin{tabular}{@{}c@{}}$mAcc$\end{tabular}\\

            \cmidrule{1-15}
            Vanilla CNN & RGB & \xmark  & 36.0 & 35.7 & 31.4 & 20.2 & 16.7 & 14.9 & 36.8 & 32.4 & 29.7 & 87.9 & 83.9 & 76.8  \\
             \cmidrule{1-15}
             \multirow{4}{*}{Partial CNN} & RGB  & \xmark & 38.6 & 37.6 & 33.5 & 22.6 & 19.0 & 16.2 & 36.0 &  32.4& 29.0 & 81.8 & 81.8 & 73.7 \\
             & RGB & \cmark & 41.2 & 40.5 & 36.2 & 19.0 & 16.7 & 13.5 & \underline{39.0}  & \underline{36.8} & \underline{32.4} & \textbf{94.9} &\textbf{91.0} & \textbf{82.6}\\
             & Convolution & \xmark & 38.6 & 37.6 &33.6 & 22.6 & 17.9 & 16.2 & 36.8 & 33.1 & 29.5 &  88.9 & 84.8 & 78.0  \\
             & Convolution & \cmark & \underline{44.4} & \underline{43.1} & \underline{38.5} & 25.0 & \underline{22.6} & \underline{19.3} & 34.6 & 33.1 & 28.8  & 88.0 & 84.8 & 77.3 \\
              \cmidrule{1-15}
             \multirow{2}{*}{\begin{tabular}{@{}c@{}c@{}} Partial CNN \\ + Object \end{tabular}} & Convolution&\xmark& 39.9 & 38.9 &	34.4 & \underline{27.4}	& \underline{22.6} & 18.9 & 31.6& 27.9& 24.7& 91.9 & 87.9 &79.7\\
               
             & Convolution  &\cmark	&\textbf{49.2} & \textbf{48.6} & \textbf{43.0} & \textbf{35.7} & \textbf{32.1} & \textbf{27.0} & \textbf{47.1} & \textbf{44.9}&	\textbf{39.8} & \underline{92.9} & \underline{88.9} & \underline{81.0}\\

\bottomrule
        \end{tabular}
    }

    \caption{Ablation studies. Results of risk object identification in four scenarios on the HDD. The unit is \%. The best and second performances are shown in bold and underlined, respectively.}
    \vspace{-15pt}
    \label{table:cause_object_identification_acc}
\end{table*}

\section{Experiments}

\subsection{Dataset}
We evaluate the proposed framework on the HDD dataset~\cite{Ramanishka_behavior_CVPR_2018}, a multisensory 104-hour naturalistic driving dataset providing a 4-layer representation of tactical driver behavior.
The \textbf{Stimulus-driven action} layer includes behaviors such as `stop' and `deviate' and the \textbf{Cause} layer denotes the reasons for behavioral changes such as an oncoming vehicle.
For example, while going straight, an oncoming vehicle causes the driver to stop.
In total, the dataset has 6 \textbf{Cause} scenarios, i.e., \textit{Stopping for Congestion}, \textit{Stopping for Crossing Vehicle}, \textit{Deviating for Parked Vehicle}, \textit{Stopping for Pedestrian}, \textit{Stopping for Sign}, and \textit{Stopping for Red Light}.
The first four scenarios are selected to evaluate the proposed risk object identification framework.
We utilize the frame-level driver behavior label (i.e., `Go' and `Stop') to train our driving model.

The dataset consists of 137 sessions, and each session represents a navigation task performed by a driver.
In \cite{Ramanishka_behavior_CVPR_2018}, the authors split the dataset into training and testing sets according to the vehicle’s geolocation.
We use the same train/test data split as in \cite{Ramanishka_behavior_CVPR_2018} (i.e., 100 sessions for training and 37 sessions for testing).
%
%
The dataset provides annotations of risk objects for a tiny portion of the dataset, making it infeasible to train a robust supervised learning-based two-class object detection model as in~\cite{Zeng_risk_cvpr2017,Gao_goal_icra2019}.
The statistics of train/val/test samples are presented in TABLE~\ref{table:dataset_statistics}.


%
We use accuracy, i.e., the number of correct predictions over the number of ground truth samples as the evaluation metric.
An accurate prediction is that an Intersection over Union (IoU) score is greater than a threshold.
Like~\cite{lin2014microsoft,zhang2019self}, accuracy at IoU thresholds of 0.5 and 0.75 are reported, as well as mean accuracy $mACC$, which is the average accuracy at 10 IoU thresholds evenly distributed from 0.5to 0.95.

\subsection{Implementation Details}
We implement our framework in PyTorch~\cite{PyTorch} and perform all experiments on a system with Nvidia Quadro RTX 6000 graphics cards.
The framework takes a sequence of frames with a resolution of $299\times299$ at 3 fps, and $T$ is set to 3 in all the experiments, approximately 1s.
The corresponding input mask maintains the same size as the input image.
The convolutional backbone is a InceptionResnet-V2~\cite{szegedy2017inception}, pre-trained on ImageNet~\cite{Russakovsky_imagenet_ijcv2015} and is modified with partial convolution layers~\cite{liu2018image,liu2018partial}.
A Detectron model~\cite{Detectron2018} trained on MSCOCO~\cite{lin2014microsoft} is used to generate object bounding boxes.
RoIAlign extracts an object representation with a size of $20\times8\times8$ from the Conv2d\_7b layer, which is then flattened into a 1280-dimensional vector.

We follow the same initialization strategy as in~\cite{Xu_TRN_iccv2019}, i.e., the hidden state units is set to 512, and dropout keep probability~\cite{srivastava2014dropout} is set to 0.5 at hidden state connections LSTM.
The aggregated feature $g$ concatenated from ego features and object features is a 1-D vector with 1024 channels.
Similar to~\cite{kim2017interpretable}, the output sizes of 3 fully-connected layers before the final binary classifier are 100, 50, and 10, respectively.
The network is trained end-to-end for 10 epochs with batch size set to 16.
We use Adam~\cite{Kingma_adam_arvix2014} optimizer with default
parameters, learning rate 0.0005, and weight decay 0.0005.

\subsection{Ablation Studies}
\vspace{-2pt}
%
%
We evaluate three aspects of our framework in TABLE~\ref{table:cause_object_identification_acc}.

\textbf{Architecture of the Driving Model}. Our proposed driving model uses features from CNN features and object features.
For CNN features, we test two backbone features, i.e., vanilla convolution and partial convolution.

\textbf{Intervention Mask.} The input to Partial CNN includes an extra mask, offering two options to intervene an image.
We either input an RGB image with selected region masked out or feed in a binary mask with the selected region set to 0.
We denote the two ways of intervention as `RGB mask' and `Convolution mask' in TABLE~\ref{table:cause_object_identification_acc}.

\textbf{Training with Intervention}.
To understand the intervention's effect on training driving models, i.e., using intervention to generate more traffic configurations, we explore two experimental settings — training with and without intervention.
Note that we always use a convolution mask to remove selected objects while using the training with the intervention strategy for Partial CNN.
In the Partial CNN + Object model, we additionally remove the selected object features during message passing.

Our final framework (last row in TABLE~\ref{table:cause_object_identification_acc}) boosts the $mACC$ by $11.6\%$, $13.5\%$, $11\%$, and $7.3\%$, respectively, compared with the lowest accuracies.
It ranks first in three scenarios (Crossing Vehicle, Crossing Pedestrian and Parked Vehicle) and second in Congestion case.
Training with intervention always leads to an increase in accuracy when object-level information is modeled.
However, it does not necessarily help the performance when the driving model is modeled using only partial CNN.

Regarding the intervened mask types, we observe that in Crossing Vehicle and Crossing Pedestrian scenarios, intervening with a convolution mask achieves higher accuracy than an RGB mask.
However, in the other two scenarios, it is the opposite.
We conjecture that, when the ego vehicle deviates for parked cars or stops for congestion, the target risk object is salient and closed to the ego vehicle.
Under such a circumstance, inputting a masked RGB frame could be sufficient for the driving model to predict correct driver behavior.
Therefore, the hallucination effect of partial convolutional layers is negligible.

\begin{table}[t]
    \small
    \centering
    \vspace{5pt}
    \resizebox{\columnwidth}{!}{
        
        \begin{tabular}
            {@{}cc@{\;}@{\;}c@{\;}@{\;}c@{\;}@{\;}c@{\;}@{\;}c@{}}
            
        \toprule
         \multirow{3}{*}{ \begin{tabular}{@{}c@{}} \\Method \end{tabular}} & \multicolumn{4}{c}{$mAcc$} \\
        \cmidrule(lr){2-5}
        & \begin{tabular}{@{}c@{}}Crossing\\Vehicle \end{tabular}
        & \begin{tabular}{@{}c@{}}Crossing\\Pedestrian \end{tabular}
        & \begin{tabular}{@{}c@{}}Parked\\Vehicle \end{tabular}
        &\begin{tabular}{@{}c@{}} Congestion \end{tabular}\\

        \cmidrule{1-5}
        Random Selection & 15.1  & 7.1 & 6.4 & 5.5 \\
        Driver's Attention Prediction *~\cite{Xia_ACCV_2018}& 16.8 & 8.9 & 10.0  &21.3 \\
        \cmidrule{1-5}
        Object-level Attention Selector *~\cite{wang2018deep}& 36.5& 21.2 & 20.1 & 8.9 \\
        Pixel-level Attention + Causality Test *~\cite{kim2017interpretable}& \underline{41.9} & \underline{21.5} & \underline{34.6}  & \underline{62.7} \\
        Ours & \textbf{43.0} & \textbf{27.0} & \textbf{39.8} & \textbf{81.0} \\
    \bottomrule
    \end{tabular}
    }

    \caption{Comparison with baseline methods. The methods with * are our re-implementation. The unit is \%. The best and second performances are shown in bold and underlined, respectively.}
    \label{table:comparison_baselines}
    \vspace{-10pt}
\end{table}

\subsection{Quantitative Evaluation}

There is no existing work being developed for risk object identification. Therefore, we re-implement three approaches~\cite{kim2017interpretable,wang2018deep,Xia_ACCV_2018}, and follow their design philosophy to select important/risk objects.
The comparison with our method is shown in TABLE~\ref{table:comparison_baselines}.

\textbf{Random Selection} and \textbf{Driver Attention Prediction}.
The results of these two methods are not directly comparable to ours, and we show the results to provide an essential measure of the difficulty of this task.
We first propose a naive baseline, randomly selecting one object as the risk object from all the detections for a given frame.
%
This method does not process any visual information.
In TABLE~\ref{table:comparison_baselines} row 2, we use a pre-trained model~\cite{Xia_ACCV_2018} to predict the driver's gaze attention maps at each frame.
We compute the average attention weight of every detected object region.
The risk object is one with the highest attention weight, indicating the driver's gaze attends this region.
The model is trained with the human gaze as supervision, which is unavailable in the HDD dataset.
Thus, we use the model pre-trained on the BDD-A dataset~\cite{Xia_ACCV_2018}.
The performance of this method is slightly better than \textbf{Random Selection}.
By visualizing the predicted attention map, we discover that the heated spots tend to cluster around the vanishing point.
%
Note that the issue has been raised in~\cite{Tawari_salient_2018}.
The reference highlights that this is one of the challenges of imitating human gaze behavior.

\begin{figure*}[t!]
\vspace{3pt}
\minipage{0.246\textwidth}
  \includegraphics[page=1,width=\linewidth]{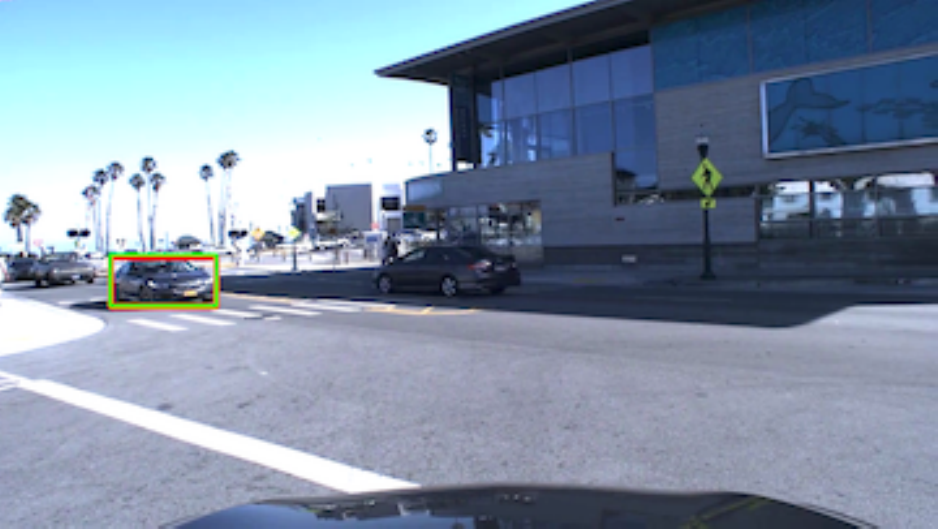}
\endminipage\hfill
\minipage{0.246\textwidth}
  \includegraphics[page=2,width=\linewidth]{figures/visualization/fig3.pdf}
\endminipage\hfill
\minipage{0.246\textwidth}
  \includegraphics[page=3,width=\linewidth]{figures/visualization/fig3.pdf}
\endminipage\hfill
\minipage{0.246\textwidth}%
  \includegraphics[page=4,width=\linewidth]{figures/visualization/fig3.pdf}
\endminipage \hfill
\vspace{4pt}
\minipage{0.246\textwidth}
  \includegraphics[page=1,width=\linewidth]{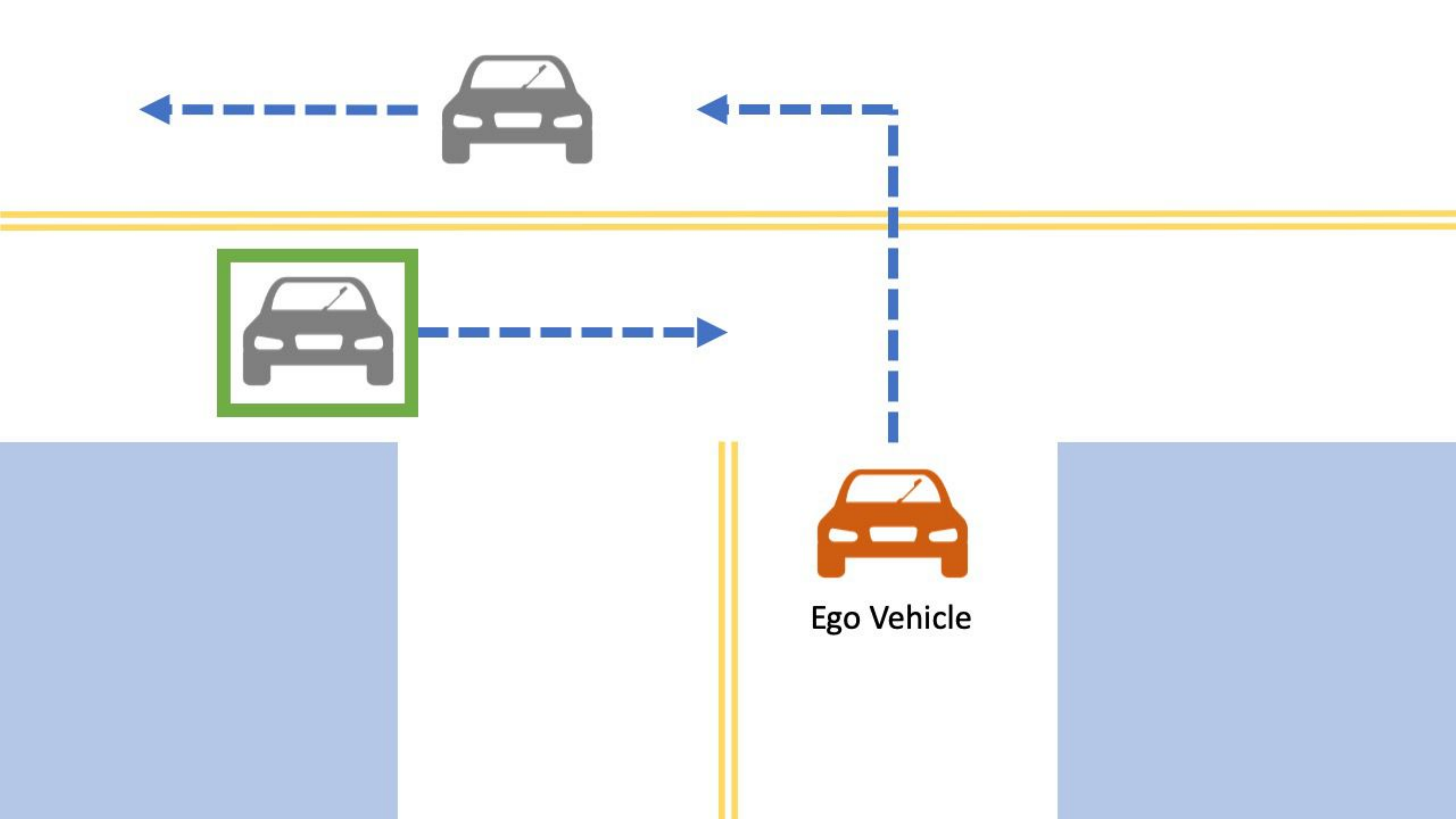}
  \caption*{(a) Crossing Vehicle}
\endminipage\hfill
\minipage{0.246\textwidth}
  \includegraphics[page=2,width=\linewidth]{figures/visualization/bev.pdf}
  \caption*{(b) Crossing Pedestrian}
\endminipage\hfill
\minipage{0.246\textwidth}
  \includegraphics[page=3,width=\linewidth]{figures/visualization/bev.pdf}
  \caption*{(c) Parked Vehicle}
\endminipage\hfill
\minipage{0.246\textwidth}%
  \includegraphics[page=4,width=\linewidth]{figures/visualization/bev.pdf}
  \caption*{(d) Congestion}
  \endminipage \hfill

\caption{Risk object identification results on sample scenarios selected from the HDD dataset. The top row shows an egocentric view where green boxes are the predicted risk objects, and ground truth ones are in red. A birds-eye-view representation is presented in the bottom row, providing information including scene layout, intentions of traffic participants, and the ego vehicle. The objects in green frames are the risk objects detected by our method.}
\label{fig:risk_object_identification}
\vspace{-8pt}
\end{figure*}

\begin{figure*}[t!]
\vspace{3pt}
\minipage{\columnwidth}
\minipage{0.55\textwidth}
  \includegraphics[page=1, width=\linewidth,trim = {0 0 0 0},clip]{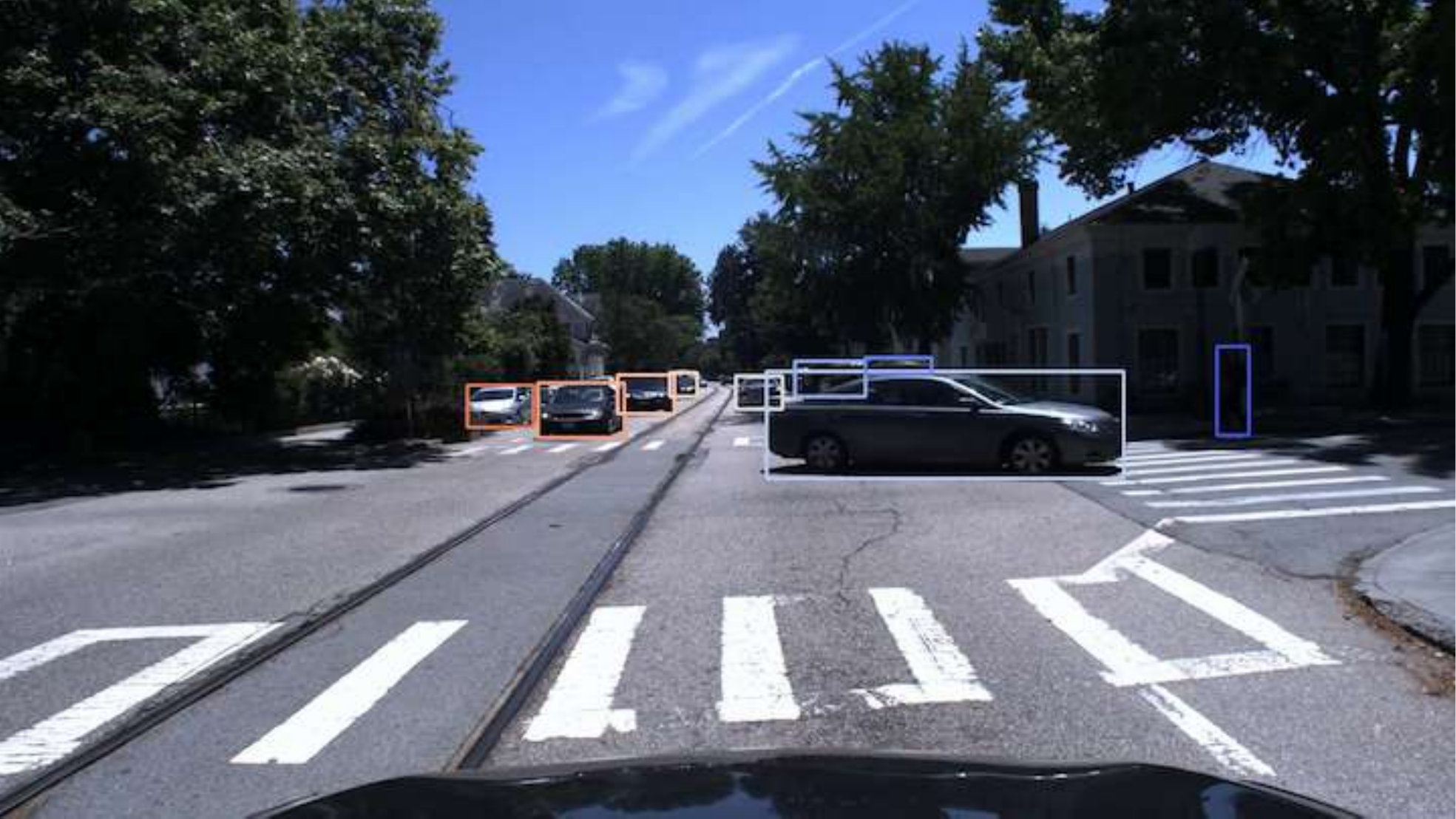}
\endminipage\hfill
\minipage{0.43\textwidth}
  \includegraphics[page=1,width=\linewidth,trim = {0 0 0 0},clip]{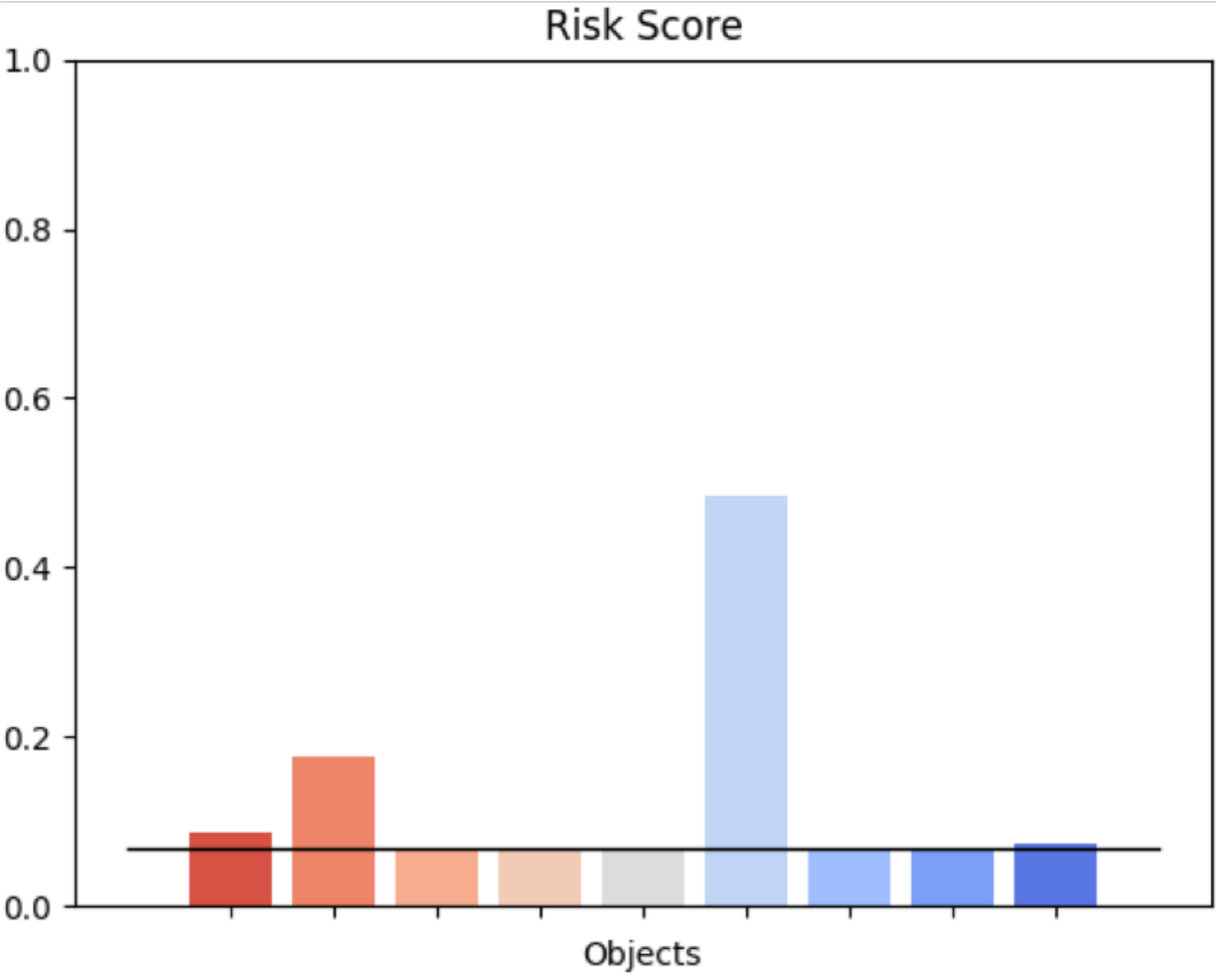}
\endminipage\hfill
\vspace{-8pt}
 \caption*{(a) Crossing Vehicle}
\endminipage\hfill
\minipage{\columnwidth}
\minipage{0.55\textwidth}
  \includegraphics[page=2, width=\linewidth,trim = {0 0 0 0},clip]{figures/riskscore/risk_score_view_compressed.pdf}
\endminipage\hfill
\minipage{0.43\textwidth}%
  \includegraphics[page=2,width=\linewidth,trim = {0 0 0 0},clip]{figures/riskscore/risk_score_bar.pdf}
\endminipage \hfill
\vspace{-8pt}
 \caption*{(b) Crossing Pedestrian}
\endminipage\hfill
\minipage{\columnwidth}
\vspace{-3pt}
\minipage{0.55\textwidth}
  \includegraphics[page=3, width=\linewidth,trim = {0 0 0 0},clip]{figures/riskscore/risk_score_view_compressed.pdf}
\endminipage\hfill
\minipage{0.43\textwidth}
  \includegraphics[page=3,width=\linewidth,trim = {0 0 0 0},clip]{figures/riskscore/risk_score_bar.pdf}
\endminipage\hfill
\vspace{-8pt}
 \caption*{(c) Parked Vehicle}
\endminipage\hfill
\minipage{\columnwidth}
\minipage{0.55\textwidth}
  \includegraphics[page=4,width=\linewidth,trim = {0 0 0 0},clip]{figures/riskscore/risk_score_view_compressed.pdf}
\endminipage\hfill
\minipage{0.43\textwidth}%
  \includegraphics[page=4,width=\linewidth,trim = {0 0 0 0},clip]{figures/riskscore/risk_score_bar.pdf}
  \endminipage \hfill
  \vspace{-8pt}
   \caption*{(d) Congestion}
\endminipage \hfill

\caption{Risk assessment results on sample scenarios selected from the HDD dataset. On the left, all detected objects are localized in colored bounding boxes. The risk score of each object is depicted in a bar chart on the right. The color of each bar is one-to-one matched to the bounding box. A black horizontal line is used to indicate the predicted `Go' score without any interventions.}
\label{fig:risk_score}
\vspace{-8pt}
\end{figure*}

\textbf{Object-level Attention Selector}.
Wang et al.~\cite{wang2018deep} designed an object-centric driving model by learning object-level attention weights, which can be further used as an object selector for identifying risk objects.
Motivated by their design, we modify the message passing in our driving model to be object-level attention and re-train our model.
We evaluate the accuracy in four scenarios based on the selected object with the highest attention weight.

\textbf{Pixel-level Attention + Causality Test}.
Kim et al.~\cite{kim2017interpretable} proposed a causality test to search for regions that influence the network's output.
They utilized the pixel-level attention map learned from an end-to-end driving model to sample particles conditioned on the attention value over an input image.
The sampled particles are clustered to produce a convex hull further to form region proposals.
Each convex hull is masked out on an RGB image, and the image is sent to the trained model to perform a causality test, iteratively.
The region, which leads to the maximum decrease of prediction performance, will be the risk object.
For a fair comparison, we replace the region proposals with object detections and utilize the pixel-level attention to filter out detections with low attention values. In the experiments, we set the threshold at 0.002.

The reason for this modification is that, compared with our detections generated from the state-of-the-art object detection algorithm, the region proposals obtained from pixel-level attention are not guaranteed to be an object, which results in an extremely low IoU and accuracy.
Additionally, the code of generating region proposals is not publicly available.
This method's performance is the closest to our results since the causality test is similar to our inference with intervention
Our performance is better because our driving model is manipulated at the object-level and is trained with the intervention strategy for robustness.

\subsection{Visualization}

We visualize the qualitative results of our method in the four scenarios.
In Fig.~\ref{fig:risk_object_identification}, ground truth risk objects are enclosed in red bounding boxes, and our predicted results are colored in green.
To better visualize the interactions between traffic participants, we provide a birds-eye-view (BEV) pictorial illustration in the second row.
The BEV figures depict the scene layout, the intention of the ego vehicle and other traffic participants', and the identified risk object in the green box.
In Fig.~\ref{fig:risk_object_identification} (b), three pedestrians are presented in the scene, crossing the road towards different destinations.
Our approach correctly tags the left-hand side pedestrian to be the risk object. 
The ego vehicle intends to take a left turn, indicating that our driving model can implicitly anticipate the ego vehicle's intention based on historical observations.

In addition to risk object identification, our framework can also assess the risk of every object in the scene.
We visualize the corresponding results in Fig.~\ref{fig:risk_score}.
%
All detected objects are localized using colored bounding boxes, and the related risk scores are in a bar chart with related colors.
The risk score of an object is defined as the predicted confidence score of `Go' when the object is masked using the proposed risk object identification framework.
A higher score of `Go' represents a higher possibility that the object influences the ego vehicle behavior.
We use a black horizontal line to indicate the predicted confidence score of `Go' without any interventions on the input.
If the score of `Go' is less than 0.5, the driver behavior is classified as `Stop.'
Our framework generates satisfactory risk assessment results qualitatively.

\begin{figure}[t!]
\vspace{3pt}
\minipage{\columnwidth}
\minipage{0.5\textwidth}
  \includegraphics[page=1,width=\linewidth]{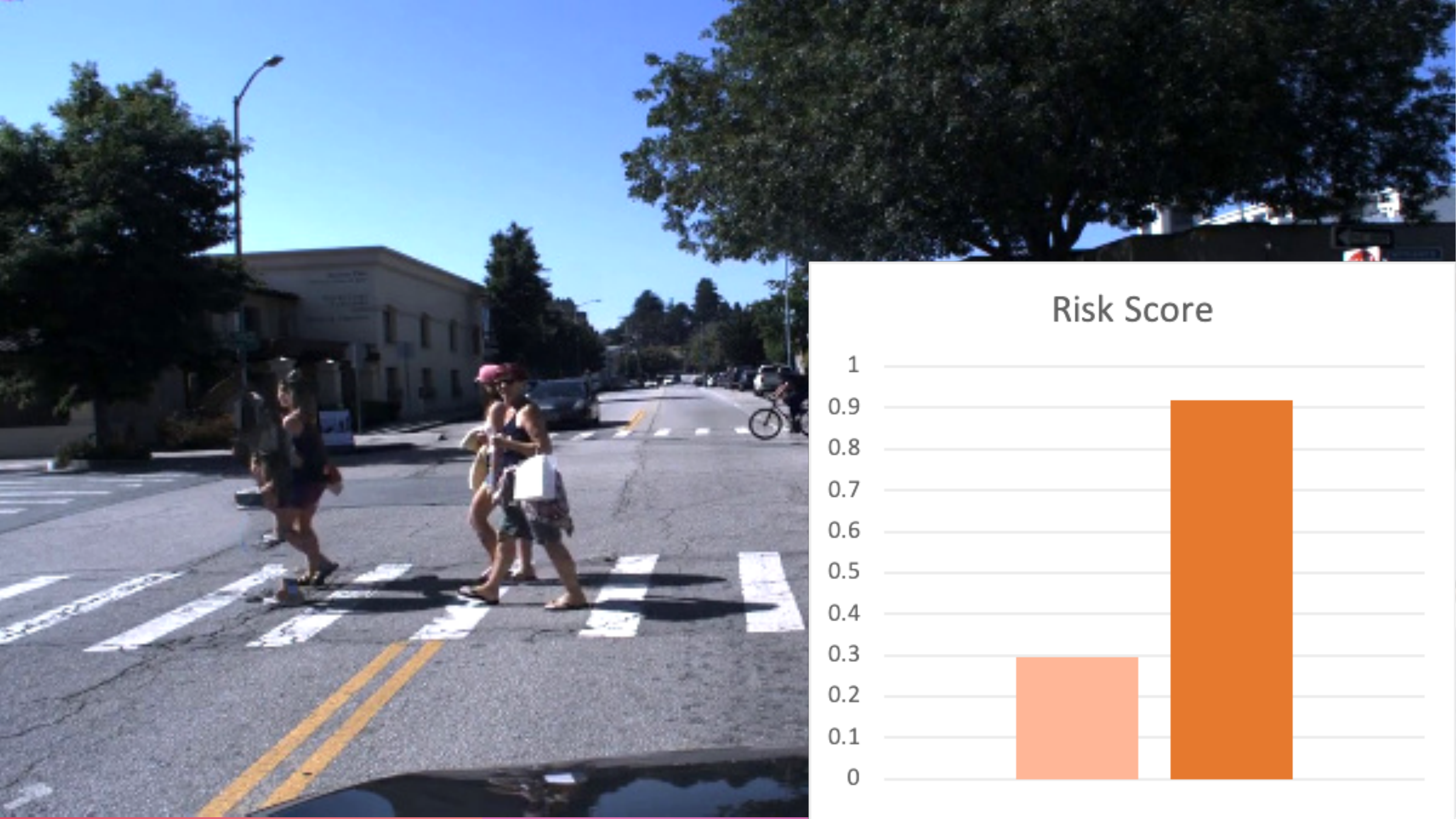}

\endminipage\hfill
\minipage{0.5\textwidth}
   \includegraphics[page=2,width=\linewidth]{figures/inpainting/inpainting.pdf}

\endminipage\hfill
\minipage{0.5\textwidth}
  \includegraphics[page=3,width=\linewidth]{figures/inpainting/inpainting.pdf}

\endminipage\hfill
\minipage{0.5\textwidth}
   \includegraphics[page=4,width=\linewidth]{figures/inpainting/inpainting.pdf}

\endminipage\hfill
\endminipage\hfill
\minipage{\columnwidth}
   \includegraphics[page=1,width=\linewidth]{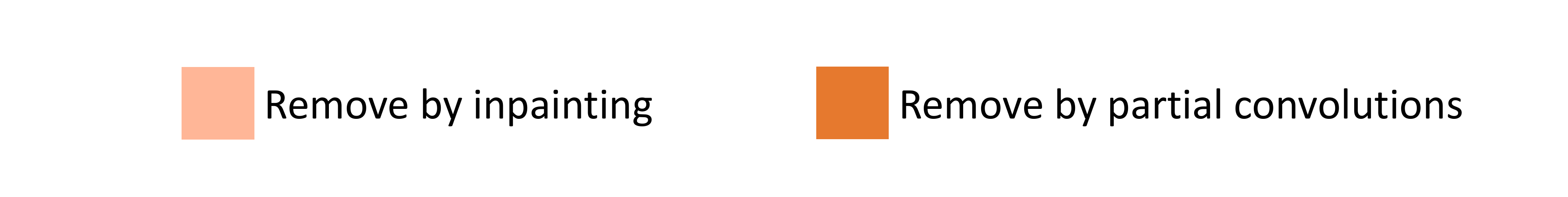}
\endminipage\hfill
\caption{An example of computed risk scores by using inpainted images and partial convolutions (our method).}
\label{fig:inpainting}
\vspace{-8pt}
\end{figure}

In Fig.~\ref{fig:risk_score} (b), when multiple risk objects (a group of people) exist, our framework assigns high-risk scores to all pedestrians.
The result seems correct at first glance.
%
However, even if one of the pedestrians was not present, the ego vehicle should have stopped.  
We conjecture that the partial convolutional operation not only hallucinates the removed area but also affects the surrounding regions due to the growing receptive field as networks go deeper.
As pedestrians are adjacent, removing one person by partial convolutions may dilute the surrounding ones and return high-risk scores.
To verify the conjecture, we manually inpainting the image by masking every pedestrian iteratively and feed the inpainted image to the same driving model without applying partial convolutions.
The corresponding results are shown in Fig.~\ref{fig:inpainting}.
A lower risk score of each pedestrian is observed that aligns with our intuition.
It also proves our conjecture that partial convolutional operations influence the behavior of the proposed driving model while more studies are needed. 
%

\subsection{Failure Cases}\label{subsec: failure}
\begin{figure}[t!]
\vspace{3pt}
\minipage{\columnwidth}
\minipage{0.5\textwidth}
  \includegraphics[width=\linewidth,trim = {0 0 0 0},clip]{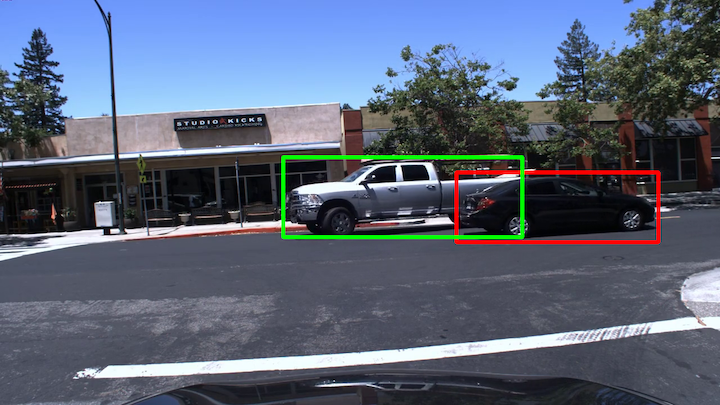}
  \caption*{(a)}
\endminipage\hfill
\minipage{0.5\textwidth}
  \includegraphics[width=\linewidth,trim = {0 0 0 0},clip]{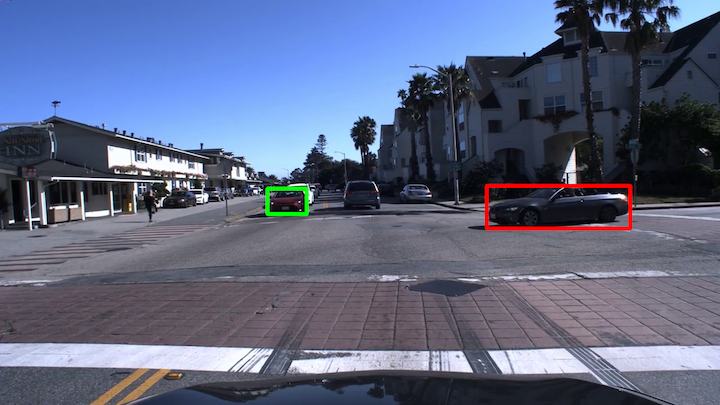}
  \caption*{(b)}
\endminipage\hfill

\endminipage\hfill
\caption{Examples of failure cases. Our prediction is in green and ground truth is in red.}
\label{fig:failure_cases}
\vspace{-8pt}
\end{figure}

While our model shows the possibility to identify the intention of the ego vehicle (Fig.~\ref{fig:risk_object_identification} (b)), there are situations that our driving model selects an incorrect risk object due to wrong intention prediction.
In Fig.~\ref{fig:failure_cases} (a), the ego vehicle plans to take a right turn and stops for the car in the red box.
However, our framework selects the white pickup truck as the risk object because of the incorrect prediction of the ego vehicle due to ambiguous and historical cues.
Additionally, in Fig.~\ref{fig:failure_cases} (b), our driving model cannot distinguish which car will move first at a 4-way stop intersection and where it is going, resulting in a wrong selection.
Hence, we believe explicitly modeling the intention of drivers, and other participants' in the driving model will render better inference results.

\section{CONCLUSIONS}
In this paper, we propose a novel driver-centric definition of risk, i.e., objects influencing drivers’ behavior are risky.
A new task called risk object identification is introduced.
We formulate the task as the cause-effect problem and propose a novel two-stage risk object identification framework based on causal inference with the proposed object-level manipulable driving model.
Favorable performance on risk object identification in comparison with strong base-lines is demonstrated on the HDD dataset.
Extensive quantitative and qualitative evaluations are conducted.
For future works, as highlighted in~\ref{subsec: failure}, explicit intention modeling of driver and traffic participants' will be beneficial.
%
%
Additionally, a more sophisticated driver behavior modeling that considers acceleration, brake, and steering are essential for reasoning the causal and effect.
Furthermore, it will be valuable for practical applications to formulate the framework into a single-stage framework, as presented in~\cite{Ke_neuralcausalmodel_arXiv2019,Nair_causalinduction_nipsw2019,Jayram_selectiveattention_nips_2019,Haan_causalconfusion_nips2019}.

\section{Acknowledgement}
The work is sponsored by Honda Research Institute USA.

\bibliographystyle{ieee}
\bibliography{reference}

\end{document}

\end{document}